\documentclass[pdflatex, sn-mathphys-num]{sn-jnl}

\usepackage{graphicx}%
\usepackage{multirow}%
\usepackage{amsmath,amssymb,amsfonts}%
\usepackage{amsthm}%
\usepackage{mathrsfs}%
\usepackage[title]{appendix}%
\usepackage{xcolor}%
\usepackage{textcomp}%
\usepackage{manyfoot}%
\usepackage{booktabs}%
\usepackage{algorithm}%
\usepackage{algorithmicx}%
\usepackage{algpseudocode}%
\usepackage{listings}%

\begin{document}

\title[Article Title]{Finding Common Ground: Using Large Language Models to Detect Agreement in Multi-Agent Decision Conferences}


\author*[1]{\fnm{Selina} \sur{Heller}}\email{sheller@rptu.de}

    \author[1]{\fnm{Mohamed} \sur{Ibrahim}}\email{mohamed.ibrahim@dfki.de}
    
    \author[1]{\fnm{David Antony} \sur{Selby}}\email{david\_antony.selby@dfki.de}
    
    \author[1]{\fnm{Sebastian} \sur{Vollmer}}\email{sebastian.vollmer@dfki.de}

\affil*[1]{\orgdiv{Data Science and its Applications}, \orgname{German Research Center for Artificial Intelligence}, \orgaddress{\street{Trippstadter Str. 122}, \city{Kaiserslautern}, \postcode{67663}, \country{Germany}}}


\abstract{Decision conferences are structured, collaborative meetings that bring together experts from various fields to address complex issues and reach a consensus on recommendations for future actions or policies. These conferences often rely on facilitated discussions to ensure productive dialogue and collective agreement. Recently, Large Language Models (LLMs) have shown significant promise in simulating real-world scenarios, particularly through collaborative multi-agent systems that mimic group interactions. In this work, we present a novel LLM-based multi-agent system designed to simulate decision conferences, specifically focusing on detecting agreement among the participant agents. To achieve this, we evaluate six distinct LLMs on two tasks: stance detection, which identifies the position an agent takes on a given issue, and stance polarity detection, which identifies the sentiment as positive, negative, or neutral. These models are further assessed within the multi-agent system to determine their effectiveness in complex simulations. Our results indicate that LLMs can reliably detect agreement even in dynamic and nuanced debates. Incorporating an agreement-detection agent within the system can also improve the efficiency of group debates and enhance the overall quality and coherence of deliberations, making them comparable to real-world decision conferences regarding outcome and decision-making. These findings demonstrate the potential for LLM-based multi-agent systems to simulate group decision-making processes. They also highlight that such systems could be instrumental in supporting decision-making with expert elicitation workshops across various domains.}




\maketitle

\section{Introduction}\label{sec1}
The development of Large Language Models (LLMs) with improved reasoning and planning capabilities has sparked significant interest in their use within multi-agent systems. Multi-agent collaboration, in particular, has been extensively studied to enhance the performance of LLMs on complex tasks beyond the capabilities of a single model \cite{guo2024large}. By enabling internal discussions between agents, these collaborative approaches seek to improve problem-solving abilities and achieve better results \cite{chan2023chateval,du2023improving,liang2023encouraging}. One application of multi-agent systems is in embodied agents, where multiple robots work together to carry out real-world tasks such as warehouse management, as well as in scientific experiments and debates \cite{guo2024large}.
This research raises new questions, particularly concerning the simulation of complex real-world scenarios. In this context, the role-playing capabilities of LLMs are utilized to simulate realistic representations of different roles and viewpoints. Key areas of application include societal simulation, gaming, psychology, recommender systems, economy, and policy-making \cite{guo2024large}. A promising application of large language model multi-agent collaborative systems is in decision conferencing and related to this also in expert elicitation workshops, which can be used for healthcare policymaking, classification of controlled substances and elicitation of informative priors for Bayesian statistical models \cite{gosling2018shelf}. 
Decision conferences are often used as a tool for addressing and resolving important issues within an organization \cite{phillips2007transparent}. Participants are guided by an impartial facilitator in a process without a fixed agenda, and without formal presentations, to debate and exchange opinions, creating a shared understanding of the issue and committing to a way forward that can be agreed upon by all participants \cite{phillips2007transparent}. Decision conferences are a form of computer-supported cooperative work (CSCW) but are inherently human-centric, with computers used only for modelling the issues as seen by the participants \cite{phillips1991decision}.
Due to their human-centric focus and specific objectives, which prioritize fostering shared understanding and a common sense of purpose, decision conferences aim to reach an agreement among all participants regarding the discussed issues \cite{phillips1991decision}. 

The field of group decision-making has been extensively studied, but there is relatively little research on decision conferences \cite{bosse2013modelling}. The increasing use of LLMs to model human social behaviour and communication raises questions about their potential for simulating decision conferences and studying communication in social groups within specific contexts \cite{chuang2024wisdom}. LLMs are often used for interaction and collaboration in social settings and simulating real-world scenarios, such as translation, question answering, essay writing, programming, and group discussions aimed at uncovering clues \cite{du2024large}. Their potential for group agreement and decision-making remains largely unexplored, despite their significant potential for application in politics and corporate decision-making. This gap leaves several questions related to the agreement in decision conferences and the usage of LLMs in multi-agent collaborative systems. Is it possible to use LLMs to detect agreement accurately? Is it then possible to use LLMs to detect agreement in open discussions of participants with different perspectives where differences in opinions are actively sought? How can a system that simulates decision conferences with open discussions and different opinions of participants be simulated using LLMs and which parts are important? Does the detection of agreement within the simulated system help the LLM agents to engage more and does it lead to a more realistic outcome of the debate?

We aim to address these questions by introducing a multi-agent collaborative decision conferencing system based on LLM agents. We focus on the primary objective of decision conferences: facilitating agreement. To achieve this, we test the ability of six different LLMs to detect agreement using benchmark datasets for stance detection and stance polarity detection, as well as evaluating the system directly through an LLM-as-a-judge approach \cite{zheng2024judging}. Additionally, we explore the role of a dedicated agent responsible for agreement detection and compare the outcomes of simulated decision conferences with and without the use of the agreement detection agent.

Our results demonstrate that LLMs can effectively perform zero-shot agreement detection across a wide range of topics and with varying contexts, making them well-suited for decision conferences that span diverse subject areas.  Additionally, we show that open-source and smaller LLMs can be applied to this task. We also find that incorporating a dedicated agent for agreement detection is crucial to ensure a detailed and conclusive debate among participating agents, leading to outcomes comparable to real-world decision conferences.

The remainder of this work is organized as follows. Section~\ref{sec2} presents a literature review, focusing on key works related to collective decision-making and multi-agent collaboration using LLMs, which are foundational to the development of the decision conference simulation system as well as the evaluation of such systems. In Section \ref{sec3}, we establish the system's foundation by detailing the structure of decision conferences and introducing the architecture for simulating decision conferences using LLMs. Section \ref{sec4} describes the evaluation methodology, detailing both objective and subjective approaches, and introduces the benchmark datasets used. In Section \ref{sec5}, we present the results of these evaluations. In contrast, Section \ref{sec6} discusses the findings concerning the task and explores their implications for other collaborative multi-agent systems that utilize LLMs for decision-making.

\section{Literature Review}\label{sec2}
Recent advancements in LLMs have significantly accelerated research in multi-agent systems and multi-agent collaboration \cite{wang2024survey}. While previous research mostly focused on fine-tuning or training agents for specific tasks in isolated environments, LLMs show new potential for application in different tasks without requiring much fine-tuning \cite{wang2024survey}. This work focuses on a specific area of real-world simulation that is often impossible without using LLMs, as traditional approaches are very preparation and modelling intensive. Although specifically simulating decision conferences with multi-agent systems remains an unexplored area, there is substantial work on related topics involving LLMs, and this work builds upon many of those ideas. Firstly, an overview of the collaborative multi-agent systems realized with LLMs is given, application areas are presented, and then the tasks of decision-making and problem-solving are considered in more detail, as well as the evaluation approaches for systems like this.

\subsection{Large Language Model Multi-Agent Collaborative Systems}
The multi-agent systems based on LLMs generally involve three communication paradigms: cooperative, debate, and competitive \cite{ijcai2024p890}. In cooperative communication, agents work together towards a common goal, while in competitive communication, agents pursue their own goals, which may conflict with those of other agents \cite{ijcai2024p890}. The debate approach, also utilized in this work, allows agents to engage in argumentative interactions and is ideal for reaching a consensus or a more refined solution \cite{ijcai2024p890}. There are four typical communication structures employed for LLM multi-agent systems: layered, decentralized, centralized, and shared message pool \cite{ijcai2024p890}. While each has its suitable application areas, this work utilizes a decentralized communication structure as it is often chosen in world simulation applications, allowing agents to communicate directly with each other \cite{ijcai2024p890}.  

\subsection{Application Areas of Large Language Model Multi-Agent Collaborative Systems}
Collaborative multi-agent systems are increasingly being used to improve the performance of LLMs in areas such as reasoning, question answering, and enhancing text quality \cite{chan2023chateval, chen2023reconcile, liang2023encouraging}. This improvement comes from breaking down complex tasks into smaller subtasks that individual agents can solve independently. To accomplish this, these systems often incorporate a planning module intended to direct agents towards more rational and dependable behaviour. The implementation of planning can vary, with some systems using feedback loops while others do not—each method offering distinct trade-offs in terms of complexity and outcomes \cite{wang2024survey}.

There is a growing interest in using these systems to simulate real-world tasks. For instance, multi-agent systems have been created to manage comprehensive business planning, involving stages such as market segmentation, customer profiling, strategy formulation, competitor analysis, and sales material creation \cite{tsao2023multi}. Similarly, multi-agent systems in software development integrate LLMs to streamline key activities, allowing for seamless natural language communication without requiring specialized models at each stage \cite{qian2023communicative}.  These systems' core is multi-agent reasoning, where agents collaborate to solve tasks. While simple debate-based approaches—where agents discuss a topic and use the resulting conclusions—can be effective, they also face challenges, such as agents getting stuck on incorrect viewpoints or failing to reach consensus, stalling progress \cite{chen2023reconcile, qian2023communicative, liang2023encouraging}.

More advanced systems address these challenges by incorporating mechanisms such as judge agents, debate moderators, or evidence management through retrieval-augmented generation to ensure more productive discussions and decisions \cite{xiong2023examining, wang2023apollo}. Additionally, some research explores the social aspects of collaborative multi-agent systems, including the use of ``social simulacra" to simulate realistic interactions within a virtual community of agents \cite{park2022social, xiao2023simulating}. Another emerging application is the simulation of international conflict resolution, where LLM-based multi-agent systems model conflict dynamics and potential resolutions \cite{hua2024warpeacewaragentlarge}.

These simulations not only assess LLM capabilities but also provide insights into the behaviour of the systems they simulate, such as identifying triggers for conflict or analyzing discourse structures in political speech \cite{hua2024warpeacewaragentlarge, wang2024survey}. Similarly, this work evaluates LLMs for their ability to detect agreement within decision-making contexts. While the primary focus is on LLM performance, the findings could contribute to a deeper understanding of decision conferences and future research in collaborative decision-making using LLM-based multi-agent systems.

\subsection{Collaborative Decision-Making}
Decision-making is a crucial human capability for everyday life. The use of Large Language Models (LLMs) to simulate decision-making processes is a new development that has expanded the possibilities for exploring complex social dynamics.
Current research focuses on making decision-making systems understandable, especially in contexts where their decisions could have real-world implications.
For example, researchers have studied how sentiment influences opinion dynamics in multi-agent systems and have identified the determinants of LLM-assisted decision-making \cite{eigner2024determinantsllmassisteddecisionmaking, ondula2024sentimental}.

Another area of research investigates the interactions among individual agents when solving complex tasks that involve decision-making, such as those related to team-building or survival tasks.
In these contexts, the dynamics of agreement and consensus-building are evaluated, particularly in relation to how these systems mirror or differ from human decision-making processes.
Studying how agents negotiate, collaborate, or conflict during task resolution provides valuable insights into the behaviour of LLM-based systems in real-world decision-making scenarios \cite{du2024large}.

\subsection{Evaluation of Large Language Model Multi-Agent Systems}
Evaluating systems composed of large language model agents is a complex challenge. Some studies use common-sense datasets or trivia questions to measure system performance \cite{wang2023apollo, liang2023encouraging, chen2023reconcile}. Another approach involves testing multi-agent systems in games, such as strategy or negotiation, to assess their collaborative or competitive performance \cite{abdelnabi2023llm}. An interesting evaluation method is presented in \cite{nardi2022graph}, where an algorithm for the axiomatic justification problem in social choice is developed. This approach depends on a structured voting framework already explored in voting theory, which differs from decision conferences that operate with more flexibility \cite{nardi2022graph}. Other approaches, such as evaluating consensus-seeking within the system \cite{chen2023multi}, focus on specific components rather than the broader system. The most promising evaluation strategies assess multi-agent systems in the context of the tasks they are designed to perform. For instance, \cite{qian2023communicative} collects statistics on code generation, while \cite{chang2024socrasynth} evaluates arguments through Socratic reasoning and formal logic principles. The Critical Inquisitive Template (CRIT) algorithm from \cite{chang2023prompting} is used across several studies to enrich evaluation \cite{chang2024socrasynth, tsao2023multi}. Additionally, comparisons between single and multi-agent systems and the use of GPT-4 in an LLM-as-a-judge approach to assessing debate dynamics further enhance the understanding of system performance \cite{chen2023reconcile, xiong2023examining}. Human evaluation remains important, as demonstrated by the involvement of experts in evaluating results \cite{tsao2023multi}.
In conclusion, evaluating LLM multi-agent systems requires a multifaceted approach, tailored to the specific goals and functions of the system. Methods that integrate task-specific performance metrics, human evaluation, and either reasoning frameworks or additional evaluation performed by an unrelated LLM hold the most promise for providing a meaningful evaluation of these complex systems.

\section{Foundations}\label{sec3}
There is a rising interest in structured and collaborative decision-making processes, especially in complex, multi-stakeholder environments as discussed in section \ref{sec2}. Previous research has investigated different methods for group decision-making using LLMs and how the results can provide insights into the process and the potential of using LLMs to simulate human group dynamics in problem-solving and general conversations. 

One way to approach collaborative decision-making is through decision conferences. In these conferences, participants engage in discussions guided by a facilitator and supported by decision analysis tools. Decision conferences provide a platform to address multi-criteria decision-making problems by combining stakeholder input and quantitative analysis, ensuring a methodical approach to solving complex issues \cite{phillips2007transparent}. There is limited research on how decision conferences function as dynamic, iterative systems and how they could be simulated. Simulation techniques can help study and optimize decision-making processes by replicating features of decision conferences, such as participant interaction, consensus-building, and the influence of facilitation techniques, within controlled, virtual environments.

In this section, we will begin by defining and explaining the fundamental principles of decision conferences to understand how they work. Then, we will introduce a framework for simulating these conferences. Our goal is to explore how decision conferences can be replicated and enhanced through simulation, ultimately improving decision-making outcomes in complex organizational contexts.

\subsection{Decision Conferences}
Decision conferences are structured meetings designed to facilitate collaborative decision-making among stakeholders. Since their first inception in 1979, these conferences have become a well-proven approach. They are utilized in various contexts, including business, politics, healthcare, and family settings, to address complex issues and reach an agreement or consensus \cite{phillips2007journal,comas2005understanding,hansen1998family}. 

Decision conferences aim to gather diverse perspectives for making informed decisions without relying on external consultants. The idea is to bring together individuals knowledgeable about the issue and provide a framework for these key players to exchange their views, provide relevant gathered data, exchange information and debate about possible solutions. This approach is effective because the information needed to resolve the issues often already resides within the participants, not necessarily in reports or other documents external consultants have access to \cite{phillips2007journal}. 

These conferences typically involve structured discussions where participants evaluate options, consider potential outcomes, and work towards an agreement \cite{salinas2022decision, phillips2007journal}. The process includes preparation, the actual meeting, and follow-up phases to ensure effective implementation of decisions  \cite{salinas2022decision}.
Related to decision conferences case are expert elicitation exercises, in which experts, often without statistical expertise, are brought together to collect opinions to derive a quantitative variable, such as a Bayesian prior distribution.
An example is the Sheffield elicitation framework \cite{gosling2018shelf}.

For the conference to be successful, the issues addressed must be real, pressing concerns, such as strategic or operational challenges, including immediate crises. Despite addressing various issues, decision conferences share common elements: the participation of key stakeholders, impartial facilitation, real-time modelling with continuous display of the evolving model, and an interactive, iterative group process. In a classical decision conference, participants are divided into participants and a facilitator \cite{phillips2007journal}.

The participants should represent the main perspectives on the issue. While it can be beneficial, they do not always need to be the decision-makers. This is often true in public sector decision conferences, where the decision-makers are elected representatives, and the conference is held to provide recommendations to them \cite{phillips2007journal}.

The facilitator or consultant is a very important component of a decision conference. The responsibility of this person is to guide the group through the process and help them agree on a path forward \cite{phillips2007transparent}. The facilitator ensures that information remains neutral, the discussion stays focused, all participants are actively engaged, and the confidentiality of the discussion is upheld. The facilitator also manages group dynamics, encourages productive dialogue, and resolves conflicts without contributing to the content to remain impartial. Common techniques used by facilitators include brainstorming, scenario analysis, and voting to ensure that all perspectives are considered in a democratic and efficient decision-making process. They also learn to maintain impartiality even though taking a leadership role in the discussion can be tempting if the facilitator is also an expert in the debated topic. Facilitators are trained to use their expertise to ask the right questions to the group instead of just providing them with a solution so that a sense of ownership can develop within the group \cite{phillips2007journal}.

Critical to gaining a sense of ownership in the group of participants is the real-time modelling of everything discussed and agreed upon, ensuring that all information is continuously available to participants for open discussion, modification, and editing. The model is built step by step, with each addition carefully explained to maintain transparency throughout the process. As the model evolves during the conference, it may prompt new perspectives, leading to revisions or rework. Dissatisfaction with certain model elements often fuels constructive debate and drives the group’s dialectic process. This dynamic nature of the conference allows for flexibility in exploring alternative solutions and adjusting decisions as necessary. The interactive, real-time evaluation of options encourages participants to engage deeply with the problem and consider the consequences of their decisions \cite{phillips2007journal}.

Decision conferences are particularly useful when addressing complex problems that require input from multiple stakeholders with potentially competing interests. These settings are common in public policy, corporate strategy, healthcare, environmental planning, and other fields where decisions have broad, long-term implications. The structured nature of a decision conference helps manage participants' cognitive load, reduce biases, and maintain clarity throughout the process \cite{comas2005understanding}. 

Despite their effectiveness, decision conferences come with challenges. Coordinating input from multiple stakeholders—especially those with diverse expertise and interests—can result in lengthy discussions and potential gridlock. Additionally, the success of the conference depends on the skill of the facilitator and the quality of the decision-support tools used \cite{comas2005understanding}. Therefore, careful planning and execution are relevant to achieving the desired outcomes. Even when the recommendations from the conference are not always implemented, participants highly value the process. This is because the shared understanding develops a sense of common purpose and helps align participants on a path forward, even if they don’t all agree on the final decisions \cite{phillips2007journal}.

Unlike typical workshops, decision conferences are designed to help decision-makers commit to a course of action by making recommendations, rather than achieving a specific technical objective. This positions decision conferences as a socio-technical approach to addressing organizational challenges \cite{phillips2007journal}. 

\subsubsection{Structure of Decision Conferences} \label{subsec32}
Establishing a well-defined structure is crucial for the successful facilitation of decision conferences, as it significantly impacts the overall quality of the outcome and the satisfaction levels of the participants. The decision conference process consists of four stages, beginning when someone in the organization identifies a need, such as a gap between desired and actual performance, the necessity for new strategies due to environmental changes, or the need for updated policies to stay relevant \cite{phillips2007journal}. 

Once a motivation for change is established, the facilitator engages the client in the first stage: preparation. During the preparation stage, the facilitator meets briefly with the decision maker and key staff members to explore the issue and determine whether a decision conference can effectively address the problem. If the issue is deemed suitable for a decision conference, the team establishes the conference's objectives and identifies the key players who should attend. Key players are individuals whose perspectives are valuable for resolving the issue, which includes those from the affected parts of the organization, as well as representatives from other areas or external experts if specialized information is needed. At the end of the preparation stage, a calling note is created and sent to participants, which includes the purpose of the decision conference, administrative details, an introduction to decision conferencing, any preparatory tasks required of the participants and also a request that participants commit to uninterrupted attendance throughout the typically two-day duration of the conference \cite{phillips2007journal}.

The decision conference itself begins with a review of the objectives, which can be discussed and adjusted by the participants to ensure agreement. This process establishes the primary task, aligns the group with a shared understanding, and provides the facilitator with a clear goal to guide future interventions. The conference proceeds through three main stages: discussing the issues, building a model of those issues, and exploring the results. Participants are encouraged to compare the outputs and stages with their own judgments and gut feelings, reporting any concerns to generate new insights. The model is refined through each iteration until a shared understanding is achieved and all perspectives are incorporated in some form. Through the shared understanding, a commitment to actions to solve the issue can be given by the participants and the decision conference concludes with the preparation of a report by the facilitator \cite{phillips2007journal}.

\subsection{Simulation of a Decision Conference with Large Language Models}
After examining the theoretical framework and stages of decision conferences, we will now introduce an approach for simulating these conferences using an LLM agent system. This simulation offers valuable insights into the decision-making capabilities of LLMs, particularly in tasks such as identifying agreements and fostering consensus in debates. To maintain relevance and comparability to real-world decision conferences, only specific stages of the process are simulated. The chosen stages for simulation are those that contribute to the collective understanding of the participants, encompassing issue discussion, model construction, result exploration, and model refinement. This approach also has the potential to provide recommendations to decision-makers who are not directly engaged in the simulation, thereby potentially enhancing their decision-making process.

\subsubsection{Architecture of the Simulated Decision Conference} 
\begin{figure}[t]
    \centering
    \includegraphics[width=\textwidth]{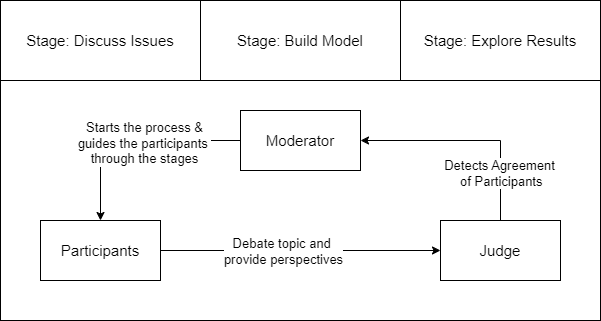}
    \caption{\label{fig:fig1}Schematic of the simulated decision conference process}
\end{figure}

To make the simulation possible a carefully designed architecture is required that mirrors the decision conference process while utilising the capabilities of the LLM agents. In Figure~\ref{fig:fig1} the simulation system's architecture is outlined, detailing how LLM agents engage in the stages detailed in section \ref{subsec32} to simulate participant interactions, model building, and consensus formation.

The system, as detailed in Figure~\ref{fig:fig1}, is initiated by providing the moderator agent with the issue for the decision conference. The moderator agent then starts the first stage, where the participant agents discuss the issue. Once each participant agent has contributed their perspective, a judge agent determines whether an agreement is reached or the participants still have something to debate. The judge agent informs the moderator agent if an agreement is reached or not, prompting the moderator to either let the participants debate more on the topic or move to the next stage: model building. This stage operates similarly to the previous one, with participant agents discussing until they agree on a model, at which point the judge agent signals the moderator to proceed to the results exploration stage. As before, participants discuss and refine the model until agreement is reached. 

Since the stages in Figure~\ref{fig:fig1} are revisited in a loop, with content sometimes evolving and sometimes staying the same, the moderator agent plays an important role in ensuring that participant agents adhere to the structure and stay focused on the task-relevant to each stage.  

The decision conference system outlined in Figure~\ref{fig:fig1} is implemented using the Autogen framework \cite{wu2023autogen}. This framework allows flexible LLM swapping and easy modification of agent definitions and interactions. To support the described procedure, a custom speaker selection function \ref{alg:speaker} must be implemented, as the default options provided by Autogen are not sufficient to manage the process effectively. 

\begin{algorithm}
\caption{Speaker Selection Function used in the Simulated Decision Conference}\label{alg:speaker}
\begin{algorithmic}
\If{$\operatorname{length}(\text{messages}) \leq 1$}
    \State Speaker $\gets$ moderator
\ElsIf{last\_speaker $=$ judge\_agent}
    \If{\textsc{Agreement} $\in$ message\_content}
        \State Speaker $\gets$ evaluation\_agent
    \ElsIf{\textsc{Debate} $\in$ message\_content}
        \State Speaker $\gets$ moderator\_agent
    \EndIf
\ElsIf{last\_speaker $=$ evaluation\_agent}   
    \State Speaker $\gets$ moderator\_agent
\ElsIf{last\_speaker $=$ moderator\_agent}   
    \State Speaker $\gets$ participant1\_agent
\ElsIf{last\_speaker $=$ participant1\_agent}   
    \State Speaker $\gets$ participant2\_agent
\ElsIf{last\_speaker $=$ participant2\_agent}   
    \State Speaker $\gets$ judge\_agent
\Else{}
    \State Speaker $\gets$ auto
\EndIf
\end{algorithmic}
\end{algorithm}

In addition to the schematic in Figure~\ref{fig:fig1}, the speaker selection function \ref{alg:speaker} includes an additional evaluator agent. This agent provides an evaluation of the debate of the participant agents by scoring them on a scale of one to ten according to factors such as: 
\begin{enumerate}
    \item Clarity: How clear is the exchange? Are the statements and responses easy to understand?
    \item Relevance: Do the responses stay on topic and contribute to the conversation's purpose?
    \item Conciseness: Is the dialogue free of unnecessary information or redundancy?
    \item Politeness: Are the participants respectful and considerate in their interaction?
    \item Engagement: Do the participants seem interested and actively involved in the dialogue?
    \item Flow: Is there a natural progression of ideas and responses? Are there awkward pauses or interruptions?
    \item Coherence: Does the dialogue make logical sense as a whole?
    \item Responsiveness: Do the participants address each other's points adequately?
    \item Language Use: Is the grammar, vocabulary, and syntax appropriate for the context of the dialogue?
    \item Emotional Intelligence: Are the participants aware of and sensitive to the emotional tone of the dialogue? 
\end{enumerate}

This evaluation of the debate is based on the LLM-as-a-judge approach which is also applied later on to assess the judge agent's performance \cite{zheng2024judging}. The evaluation of the participant agents while the system is running provides a preliminary indication of how well the decision conference proceeds and whether it holds value for further evaluation. This reduces the manual effort required to review long transcripts and conversations.

The speaker selection function \ref{alg:speaker} and the general architecture of the system displayed in Figure~\ref{fig:fig1} allow the simulation to be easily extended, for example by adding more participant agents. This flexibility enables the simulation of decision conferences with larger, more diverse groups and the incorporation of different personas or perspectives. 

To evaluate the judge agent in this work, only two participants are required to assess the ability of LLMs to detect agreement. It is important to first determine whether the judge agent can reliably recognize agreement, as forms the basis for more complex evaluations and ensures that the system can effectively mediate debates. Future evaluations of the system could explore the impact of scaling the number of participants, integrating more complex personas for each participant, or handling topics that are more difficult to reach an agreement on. These enhancements would further challenge the judge agent and provide deeper insights into its performance in more dynamic discussions.

\section{Methodology}\label{sec4}
Having established the theoretical foundations of decision conferences and described the architectural framework for simulating decision conferences with LLMs, we now want to address and explain the methodology for evaluating the effectiveness of judge agents in this context. In this section, a two-part evaluation strategy is presented: objective evaluation using benchmark datasets and subjective evaluation using an LLM-as-judge approach alongside manual evaluation \cite{zheng2024judging}.

The objective evaluation involves a comprehensive examination of the judge agent's ability to recognize agreement using established benchmark datasets. This quantitative approach provides a basis for evaluating performance metrics and the robustness of LLMs under controlled conditions.

This is complemented by the subjective evaluation of LLMs in the role of judge, where further linguistic understanding and contextual reasoning come into the picture. This evaluation examines the adaptability of LLMs in decision conference scenarios and the consistency of their decisions as well as the need to use LLMs in this scenario.

\subsection{Objective Evaluation of Large Language Models as Judge Agent}
The objective evaluation focuses on quantitatively assessing the judge agent's ability to accurately detect agreement between debating agents. This evaluation is conducted using established benchmark datasets designed to measure two aspects of agreement detection, stance detection and stance polarity detection. 

Stance detection is an important task in natural language processing and involves determining the position or attitude of a speaker or writer towards a specific topic or proposition \cite{gul2024stance}. It is mostly used to analyze social media content to gain insights into public sentiment and societal trends \cite{gul2024stance}. When it comes to detecting agreement, understanding the stance can help determine whether the speaker or writer agrees with previously stated propositions or ideas. If the stance aligns with the initial statement, it suggests agreement. On the other hand, if the stance is opposing, it indicates disagreement. LLMs are particularly appealing for stance detection because fine-tuning traditional machine learning models, such as support vector machines or deep learning models, can be both time-consuming and resource-intensive \cite{gul2024stance}. LLMs can be guided through prompting and offer a more efficient and flexible approach, eliminating the need for extensive fine-tuning while allowing the models to handle a wide variety of content that would have been difficult to capture with conventional methods \cite{gul2024stance}. 

Stance polarity refers to the direction of the stance, which means whether it is positive (supportive or negative (opposing). This is directly related to agreement detection as positive polarity can suggest agreement, while negative polarity can suggest disagreement. In a debate understanding the polarity of a stance can allow the system to classify expressed sentiment as agreeing or disagreeing with the prior statements or positions. 

Both stance detection and stance polarity detection are well-established tasks in natural language processing (NLP), and many benchmark datasets have been developed to evaluate models on these tasks. Most of these datasets are sourced from social media platforms like X (formerly Twitter) and Reddit, where diverse opinions and stances are readily available. Among the most well-known are the SemEval-2016 Stance Dataset and the P-Stance Dataset, which have been widely used for assessing model performance in the task of stance detection for social media \cite{mohammad-etal-2016-semeval, li-etal-2021-p}.

While these datasets are useful for evaluating stance detection in short, opinionated social media posts, they are not ideally suited for detecting agreement in structured debates. Debates typically involve extended arguments and require an understanding of discourse dynamics that go beyond social media. More specialized datasets that focus on conversational or argumentative agreement are necessary to evaluate LLMs in the context of debates, as these better reflect the complexities of formal discussion environments. The two datasets used in this work to evaluate an LLM agent in the task of stance detection are the Varied Stance Topic Data (VAST) Dataset and the Claim Stance Classification Dataset \cite{allaway-mckeown-2020-zero, bar-haim-etal-2017-stance}. Both capture a wider range of topics and linguistic variation than typical social media datasets, making them more suitable for evaluating zero-shot approaches to stance detection and stance polarity detection. In contrast to social media-based datasets, which often contain short, informal texts with limited context, these datasets offer more complex and diverse linguistic structures that provide a better testing ground for the capabilities of LLMs. They are also used in other studies exploring the potential of LLMs in computational argumentation and other related tasks \cite{chen2023exploring, ubaida2024can}.

Given that the two datasets employ different metrics for evaluating model performance, we use a set of metrics containing accuracy, precision, recall and $F_1$~score to provide an extensive overview of the capabilities of LLMs for stance detection in both of the datasets and stance polarity detection in the Claim Stance Classification Dataset.

\subsubsection{Description of the Datasets}
The test set of the VAST dataset contains 676 examples annotated for the task of stance detection. The samples contained in the dataset are newly annotated examples from a part of the Argument Reasoning Comprehension (ARC) Corpus \cite{allaway-mckeown-2020-zero}. The labels for stance detection include pro, con and neutral and crowdsourcing was used to annotate the samples. 

The test set of the Claim Stance Classification Dataset contains 1.355 examples annotated for the tasks of stance detection and stance polarity detection. The topics were selected from the debate motions database and the claims were then manually collected from Wikipedia articles \cite{bar-haim-etal-2017-stance}. The labels for stance detection include only pro and con while the labels for stance polarity detection include positive, negative and neutral. 

\subsection{Subjective Evaluation of Large Language Models as Judge Agent}
The subjective evaluation aims to assess the judge agent's performance in detecting agreement through a more qualitative lens, incorporating human preferences and interpretative reasoning. This evaluation involves using an LLM-as-a-judge approach to evaluate how well the agent can detect agreement between participants in simulated decision conferences. By using this approach, we expand the objective evaluation by the addition of the understanding of context, tone, and subtle agreement that may be difficult to quantify through objective metrics alone.

There are several variations of the LLM-as-a-judge approach, such as pairwise comparison, single-answer grading, and reference-guided grading, each offering distinct advantages and drawbacks \cite{zheng2024judging}. In the pairwise comparison method, where the LLM is presented with a question and two answers, scalability becomes a challenge as the number of possible answer pairs increases with each additional participant \cite{zheng2024judging}. On the other hand, single-answer grading, where the LLM directly assigns a score to a single response, may struggle to capture subtle nuances between closely related answers \cite{zheng2024judging}. This method can also be influenced by the underlying model, leading to instability in results if the model is updated or changed \cite{zheng2024judging}. Reference-guided grading, which evaluates responses using a predefined reference answer, includes a lot of human work for our task as there is no reference answer available for this specific problem \cite{zheng2024judging}.

One of the benefits of using an LLM-as-a-judge approach is explainability \cite{zheng2024judging}. LLMs provide explanations for the scores or decisions they make, offering interpretable outputs for each judgement they make \cite{zheng2024judging}. This ability to trace the decision-making process can be particularly helpful to ensure transparency, especially in complex scenarios like debates or decision conferences, where understanding the reasoning behind an evaluation is very important.

Another major advantage is the reduction of human involvement in the evaluation process \cite{zheng2024judging}. By automating the grading or judging task, LLMs minimize the need for human intervention, reducing the time required for manual assessment \cite{zheng2024judging}. This scalability allows for more evaluations across a larger number of responses or participants, without sacrificing the quality of judgment \cite{zheng2024judging}.

While the LLM-as-a-judge approach offers a wide range of applications—from writing, roleplay, extraction, reasoning, math and coding to reasoning tasks and multi-turn questions, its flexibility and adaptability also make it suited to the needs of evaluating the judge agent in the decision conferencing system proposed in section \ref{sec3} \cite{zheng2024judging}.

To enable the LLM-as-a-judge agent to evaluate the output of the decision conferencing system, a realistic decision conference simulation is required. Since real-world reports from decision conferences typically summarize only the final results and lack detailed information about the discussions, it is essential to use a comprehensive real-world example that provides more context. A decision conference focused on formulating and evaluating drug policy offers extensive details on the structure, debated topics, and outcomes, making it a suitable reference for comparison with the simulated system \cite{rogeberg2018new}. For a clear evaluation of the system, the focus should only be on the stages of the decision conference where the debates between participants happen and that are detailed in the architecture of the system in Figure~\ref{fig:fig1}. We assume that the other stages, namely recognizing the need and deciding on the goals to achieve, have already taken place and take the results of these stages from the real decision conference. The topics that the real stakeholders have decided to debate on are: 
\begin{enumerate}
    \item develop a set of criteria for assessing drug policy outcomes
    \item define a set of generic drug regulatory regimes to encompass a broad spectrum of general approaches to controlling drugs that are deployed in practice
    \item apply the model to alcohol
\end{enumerate}
The need is given because a decision conference with real people has already taken place.

The decision conferencing system described in section \ref{sec3} is used to simulate a structured debate among participant agents on these three interconnected topics, with the discussion guided by a moderator agent. The judgements of the judge agent regarding the agreement of the participants is evaluated at the end of the debate using the single answer grading of the LLM-as-a-judge approach which is described above. To adapt this process to the context of our evaluation, the LLM is given a prompt containing specific details. The model is instructed to assess factors such as stance detection, stance polarity, and overall sentiment to determine the level of agreement between the participant agents. This assessment of the debate should then be compared to the outcome of the judge agent assessment. Since the judge agent is instructed to only detect if an agreement is there or not and the format of the answer is very important for the remaining parts of the system to work, the format of the output and its adherence to a standard should also be evaluated. Additionally, a brief explanation of the evaluation should be provided as well as the rating of the LLM used as the judge agent on a scale from zero to ten.

Another small part of the subjective evaluation is the usefulness of the judge agent detecting agreement for the whole decision conference simulation system. To evaluate this a short manual evaluation compares the outcome of a simulated decision conference with and without the judge agent to the outcome of the real decision conference detailed in section \ref{sec3}. This evaluation provides valuable insight into the inner workings of the system and helps to emphasise the importance of agreement detection for the whole system. 

With the system and evaluation methodology established, we now move to the evaluation of the judge agent's performance within the decision conferencing system. The evaluation is divided into two parts, an objective evaluation followed by a subjective analysis, each providing insights into the accuracy and quality of the judge agent’s decisions. The subjective evaluation also features a short manual evaluation of why the judge agent is needed in the decision conferencing system and why it can't be omitted to achieve a good result.

\section{Evaluation}\label{sec5}
In this section, we evaluate the performance of six LLMs in the role of a judge agent within the decision conferencing system. The models assessed include Gemma 2 9B \cite{team2024gemma2}, Gemma 7B \cite{team2024gemma}, Mixtral 8x7B \cite{jiang2024mixtral}, LLaMA 3 70B \cite{dubey2024llama}, ChatGPT 3.5 Turbo \cite{ChatGPT3.5Turbo}, and ChatGPT 4 \cite{chatGPT4}. Each model is evaluated using objective and subjective evaluation as described in section \ref{sec4} to determine its effectiveness in detecting agreement and providing accurate judgments. By comparing these models, we aim to identify which LLM performs most reliably and consistently for agreement detection and subsequently as a judge agent within the decision-conferencing framework. Additionally, we explore whether open-source and smaller models can effectively handle this task, offering a more cost-efficient, accessible, and adaptable solution for wider implementation.

\subsection{Objective Evaluation of Large Language Models used as the Judge Agent}
To begin the evaluation, we first focus on the objective performance of each model. This assessment is based on quantitative metrics that measure the accuracy, consistency, and efficiency of the LLMs in stance detection and stance polarity detection on the benchmark datasets introduced in section \ref{sec4}. The datasets used in this evaluation vary significantly. The VAST dataset focuses on detecting the stance of short posts in an online debate, with minimal context about the topic or preceding posts \cite{allaway-mckeown-2020-zero}. In contrast, the Claim Stance Classification Dataset provides a standardized sentence for each topic along with shorter texts to be evaluated for stance detection and polarity \cite{bar-haim-etal-2017-stance}. This contrast offers insights into the capabilities of LLMs across different types of texts for the same task, helping us understand the extent of contextual information required for LLMs to perform effectively.

\subsubsection{Stance Detection Evaluation on VAST Dataset}
With the datasets outlined, we first evaluate the LLMs using the VAST dataset. This dataset, characterized by short posts with limited context, allows us to assess model performance in stance detection when contextual information is minimal \cite{allaway-mckeown-2020-zero}. Evaluating LLMs on this dataset poses challenges due to the class distribution in the test set: the pro and con classes are relatively balanced with 349 and 324 examples respectively, while the neutral class is underrepresented with only 2 examples \cite{allaway-mckeown-2020-zero}. This imbalance affects the metrics such as accuracy, precision, and recall, making them less reliable for this dataset. The original work introducing the VAST dataset focused solely on macro-averaged $F_1$~scores for evaluation, and we will adopt this approach to compare the performance of LLMs with that of classic models trained on the same dataset \cite{allaway-mckeown-2020-zero}. But we also take a look at the other metrics to gain a better understanding of the performance of the LLMs on the dataset even without direct comparison to classical approaches. 

\begin{table}[t]
    \caption{Evaluation on VAST Dataset with metrics accuracy, precision, recall and $F_1$~score (macro averaged)}
    \label{table:1}
    \setlength{\tabcolsep}{3pt}
    \begin{tabular}{ lrrrr }
        \toprule
       Model & accuracy & precision & recall & $F_1$~score\\
        \midrule
        Gemma 2 9B & 0.642 & 0.453 & 0.429 & 0.440\\
        Mixtral 8x7B & 0.597 & 0.468 & 0.564 & 0.434\\
        Gemma 7B & 0.430 & 0.380 & 0.459 & 0.299\\
        Llama 3 70B & \textbf{0.667} & \textbf{0.470} & 0.445 & \textbf{0.457}\\
        ChatGPT 3.5 Turbo & 0.567 & 0.429 & 0.546 & 0.408\\
        ChatGPT 4 & 0.606 & 0.468 & \textbf{0.569} & 0.437\\
        \bottomrule
    \end{tabular}
\end{table}

As shown in Table~\ref{table:1}, the accuracy of various LLMs remains competitive even with zero-shot execution using a basic prompt, without extensive prompt engineering. The top three models in terms of accuracy are Llama 3 70B, ChatGPT 4, and Gemma 2 9B. Interestingly, this lineup includes two open models and one relatively small model. But, as previously mentioned, accuracy alone is not the most appropriate metric for this dataset, and precision and recall should also be considered.

The metric considering all of these is the $F_1$~score, which is also the metric chosen in the VAST work for evaluation \cite{allaway-mckeown-2020-zero}. In Table~\ref{table:1}, it can be observed that the top three models in terms of accuracy also rank highest in the $F_1$~score. When comparing these results to the findings in the original work it is evident that the $F_1$~score of the Llama 3 70B model surpasses the performance of six of the trained models from the original study, with only two trained models performing better \cite{allaway-mckeown-2020-zero}. The top three LLM models outperform several classical machine learning algorithms, including majority class computation (CMaj) with an $F_1$~score of 0.286, bag-of-words vectors (BoWV) with an $F_1$~score of 0.349, feed-forward networks (C-FFNN) with an $F_1$~score of 0.417, and a bidirectional encoder (BiCond) with an $F_1$~score of 0.428, which was a state-of-the-art model for cross-target Twitter stance detection in 2020 \cite{allaway-mckeown-2020-zero}. Most of the LLMs evaluated in our work demonstrate superior performance compared to the C-FFNN model from the original work. The only models from the original dataset work that remain unbeaten are Bert-joint with an $F_1$~score of 0.661 and TGA NET with an $F_1$~score of 0.666 \cite{allaway-mckeown-2020-zero}. These models are highly specialized, requiring extensive adaptation to the dataset’s specifics and significant human intervention and training \cite{allaway-mckeown-2020-zero}.

While the overall performance of the top LLM models is promising compared to classical algorithms, it is important to investigate further how these models perform across the different classes. A broad metric such as $F_1$~score over all classes provides a general view, but a class-wise evaluation offers more insights into the models' strengths and weaknesses. By examining the $F_1$~scores for each class—pro, con, and neutral we can better understand how well these models handle the specific challenges of each stance category. The class-wise comparison of the $F_1$~score for the LLMs can be found in Table~\ref{table:2}.

\begin{table}[t]
    \caption{Evaluation on VAST Dataset with class-wise $F_1$~score}
    \label{table:2}
    \setlength{\tabcolsep}{3pt}
    \begin{tabular}{ lrrr }
        \toprule
       Model & Pro & Con & Neutral \\
        \midrule
        Gemma 2 9B & 0.648 & 0.674 & 0.0\\
        Mixtral 8x7B & 0.611 & 0.670 & 0.019\\
        Gemma 7B & 0.568 & 0.315 & 0.016\\
        Llama 3 70B & \textbf{0.661} & \textbf{0.709} & 0.0\\
        ChatGPT 3.5 Turbo & 0.602 & 0.593 & \textbf{0.028}\\
        ChatGPT 4 & 0.607 & 0.683 & 0.022 \\
        \bottomrule
    \end{tabular}
\end{table}

In comparison to the original work of the VAST dataset, the $F_1$~scores for the LLMs in Table~\ref{table:2} show significantly higher performance for the pro and con classes in most cases. This improvement highlights the effectiveness of LLMs in these categories compared to the classical models. However, the overall $F_1$~score for the LLMs is affected by their poor performance in the neutral class, which is rarely predicted correctly. 

The highest scoring model in the work of the VAST dataset has an $F_1$~score of 0.554 for the pro class and an $F_1$~score of 0.585 for the con class \cite{allaway-mckeown-2020-zero}. Even the least performing LLM, Gemma 7B, demonstrates a higher $F_1$~score for the pro class as can be seen in Table~\ref{table:2} compared to the models trained in the original study. Additionally, it can be seen in Table~\ref{table:2}, that five of the evaluated LLMs outperform the best model from the original work for the con class and only the smallest model falls behind. 

The evaluation of LLMs on the VAST dataset demonstrates that these models are effective for stance detection even in low-context scenarios. The ability of LLMs to accurately predict the pro and con classes is particularly significant, as these predictions are crucial for determining overall agreement. Given that the neutral class is poorly represented in the dataset, with only 2 examples in the test set, the not-really favourable performance of this class can be considered less impactful. The primary focus should remain on the LLMs' proficiency in predicting the pro and con classes, which aligns with their intended application for agreement detection.

\subsubsection{Stance Detection Evaluation on Claim Stance Classification Dataset}
 After evaluating the VAST dataset, we now focus on the Claim Stance Classification Dataset \cite{bar-haim-etal-2017-stance}. This dataset offers a more structured environment for stance detection, allowing us to assess how LLMs perform when provided with standardized topic information and shorter text excerpts \cite{bar-haim-etal-2017-stance}. The dataset maintains a relatively balanced distribution between the ``pro" and ``con" labels, with 55.3\% labelled as pro and 44.7\% as con \cite{bar-haim-etal-2017-stance}. These labels are manually annotated through a meticulous process where annotators consider the claim’s sentiment, target, and contrast relations in their assessments \cite{bar-haim-etal-2017-stance}. Given the requirement of gathering five annotations per example and clustering them to determine a majority label, the process is labour-intensive but results in a high-quality dataset suitable for evaluating various argumentation-related tasks \cite{bar-haim-etal-2017-stance}.

The work introducing the Claim Stance Classification Dataset used macro-averaged accuracy as the primary evaluation metric \cite{bar-haim-etal-2017-stance}. This metric is appropriate for the dataset, given its balanced binary classification structure. While later studies have also evaluated the dataset using $F_1$~scores, we continue to use accuracy to compare the performance of the LLMs with the baseline models from the original work \cite{chen2023exploring}. The original work explored several configurations, with different parts building up upon each other and incorporating different techniques, with the baselines representing established benchmarks in stance classification like Unigrams SVM and Unigrams+Sentiment SVM \cite{bar-haim-etal-2017-stance}. The system introduced in the work of the dataset was specifically trained on this dataset, with its components evaluated separately before being integrated into a full system \cite{bar-haim-etal-2017-stance}. As the evaluations done with other models in the following years demonstrate, surpassing the accuracy of the original system remains a challenging task \cite{chen2023exploring}.

Table~\ref{table:3} presents the results of our evaluation of the same six LLMs used in the assessment of the models on the VAST dataset. Interestingly, all models perform exceptionally well on the Claim Stance Classification dataset, which is surprising given that earlier evaluations showed models like LLaMA-2-7B and LLaMA-2-13B performed poorly on this dataset \cite{chen2023exploring}. This improvement may be attributed to advancements in better model architectures, or a broader contextual understanding of the LLMs compared to earlier iterations. These results suggest that modern LLMs have made significant steps in handling stance detection tasks, even in domains where previous models struggled.

\begin{table}[t]
    \caption{Evaluation on Claim Stance Classification Dataset for stance detection with metrics accuracy, precision, recall and $F_1$~score (macro averaged)}
    \label{table:3}
    \setlength{\tabcolsep}{3pt}
    \begin{tabular}{ lrrrr }
        \toprule
       Model & accuracy & precision & recall & $F_1$~score\\
        \midrule
        Gemma 2 9B & 0.861 & 0.860 & 0.861 & 0.861\\
        Mixtral 8x7B & 0.759 & 0.831 & 0.766 & 0.748\\
        Gemma 7B & 0.734 & 0.735 & 0.732 & 0.732\\
        Llama 3 70B & \textbf{0.934} & \textbf{0.934} & \textbf{0.934} & \textbf{0.934}\\
        ChatGPT 3.5 Turbo & 0.797 & 0.798 & 0.795 & 0.796\\
        ChatGPT 4 & 0.890 & 0.901 & 0.893 & 0.890\\
        \bottomrule
    \end{tabular}
\end{table}

In our evaluation of newer LLMs, the state-of-the-art baseline accuracy of 0.849 reached in the benchmark dataset work is surpassed by the three top-performing models: LLaMA 3 70B, Gemma 2 9B, and ChatGPT 4 \cite{bar-haim-etal-2017-stance}. Their $F_1$~scores are also relatively consistently high, significantly exceeding those from earlier evaluations of other LLMs \cite{chen2023exploring}. This indicates that progress is made regarding the accuracy of LLMs on stance detection tasks, which also scales their ability to handle argumentation tasks with greater precision.

The evaluation of the LLMs on the Claim Stance Classification Dataset demonstrates the ability of especially the top three models to not only perform well in low-context scenarios but also outperform traditional models in more complex stance detection tasks. The best-performing model introduced in the work of the benchmark dataset utilizes additional features such as sentiment and contrast detection to aid in the task of stance classification \cite{bar-haim-etal-2017-stance}. In contrast, the LLMs achieve comparable results without relying on these specialized features, working purely in a zero-shot manner with basic prompts. This emphasises the possibility of using LLMs in handling stance detection with minimal task-specific engineering. 

\subsubsection{Stance Polarity Detection Evaluation on Claim Stance Classification Dataset}
\begin{figure}[t]
    \centering
    \includegraphics[width=\textwidth]{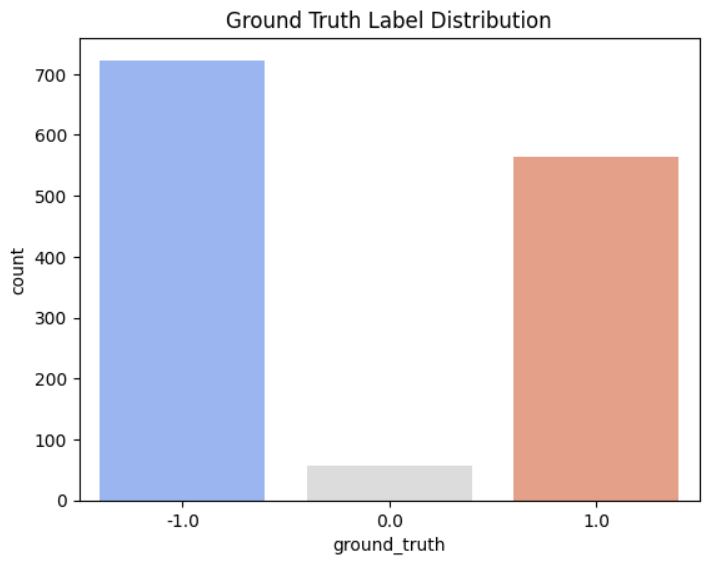}
    \caption{\label{fig:fig2} Distribution of ground truth labels in the Claim Stance Classification Dataset for Stance Polarity Detection}
\end{figure} 

\begin{table}[t]
    \caption{Evaluation on Claim Stance Classification Dataset for stance polarity detection with metrics accuracy, precision, recall and $F_1$~score (macro averaged)}
    \label{table:4}
    \setlength{\tabcolsep}{3pt}
    \begin{tabular}{ lrrrr }
        \toprule
       Model & accuracy & precision & recall & $F_1$~score\\
        \midrule
        Gemma 2 9B & \textbf{0.759} & \textbf{0.604} & \textbf{0.590} & \textbf{0.594}\\
        Mixtral 8x7B & 0.699 & 0.580 & 0.577 & 0.539\\
        Gemma 7B & 0.711 & 0.596 & 0.485 & 0.482\\
        Llama 3 70B & 0.680 & 0.527 & 0.525 & 0.524\\
        ChatGPT 3.5 Turbo & 0.730 & 0.599 & 0.576 & 0.553\\
        ChatGPT 4 & 0.725 & 0.592 & 0.570 & 0.558\\
        \bottomrule
    \end{tabular}
\end{table}

After completing the evaluation of stance detection on the VAST and Claim Stance Classification datasets, we now assess the LLMs some more with the Claim Stance Classification dataset for the task of stance polarity detection. This dataset again provides a structured context, allowing us to examine how LLMs perform in detecting stance polarity across standardized topic information and shorter text excerpts. The dataset includes three classes: positive (1), negative (-1), and neutral (0), which are, as shown in Figure~\ref{fig:fig2}, highly unbalanced \cite{bar-haim-etal-2017-stance}. Due to this imbalance, accuracy again may not be the most reliable metric for evaluation. Since the original work uses accuracy, we will first compare the LLMs' accuracy scores on this task to the baseline established in the original work of the dataset before further evaluating their performance using other metrics. Stance polarity is represented as the sentiment score baseline in the original work since it is only a part of another evaluation and not necessarily one of its own \cite{bar-haim-etal-2017-stance}.

\begin{figure}[t]
    \centering
    \includegraphics[width=\textwidth]{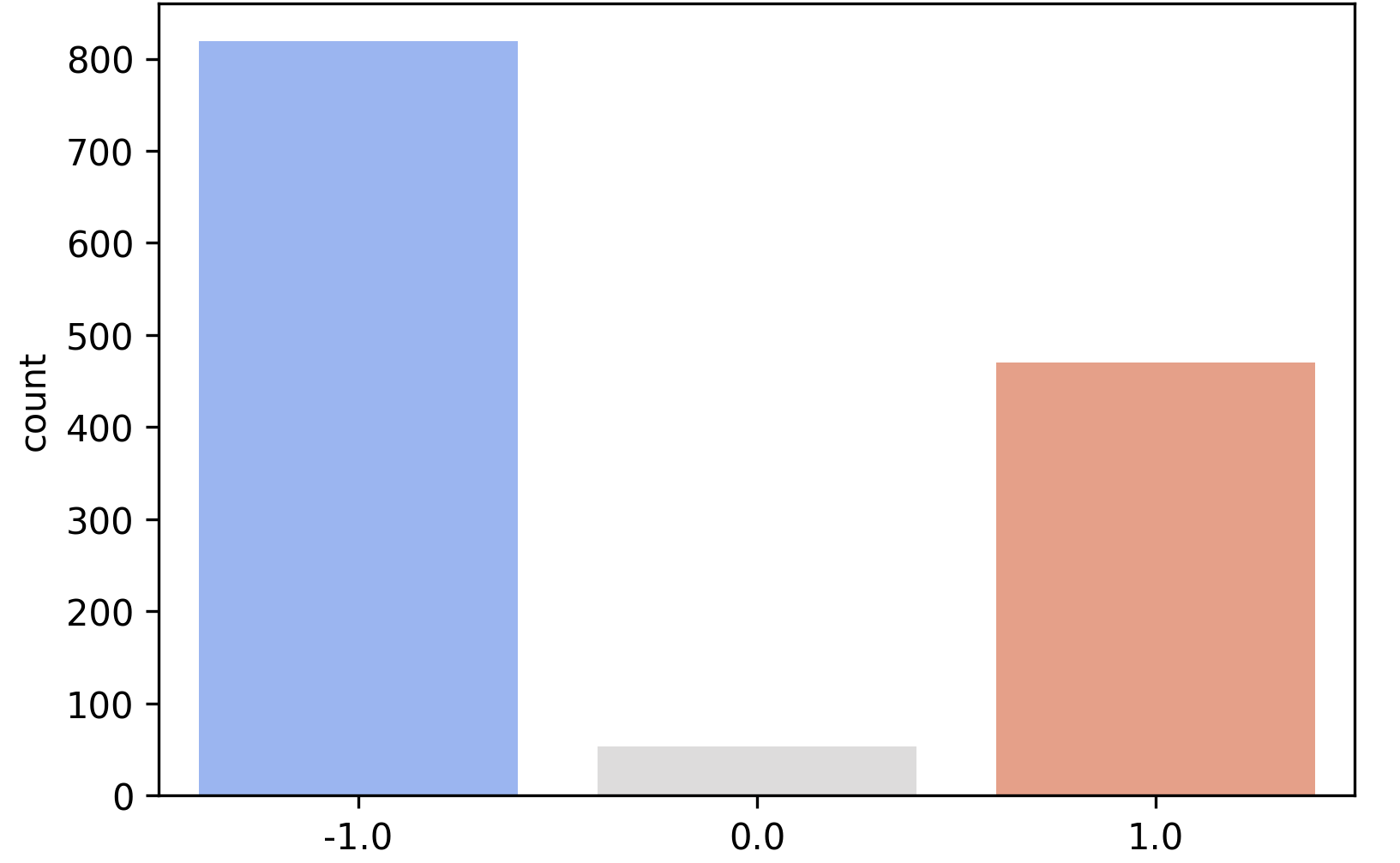}
    \caption{\label{fig:fig3} Example class distribution of Model Gemma 2 9B on the task of Stance Polarity Detection evaluated on the Claim Stance Classification Dataset}
\end{figure} 

Table~\ref{table:4} presents the results of our evaluation of the same six LLMs used in previous assessments. Only one LLM surpasses the baseline sentiment accuracy of 0.752 of the work for the benchmark dataset \cite{bar-haim-etal-2017-stance}. Gemma 2 9B is the best-performing model across all metrics, outperforming even much larger models. This highlights Gemma 2 9B's efficiency and effectiveness in stance polarity detection, demonstrating that model size alone is not always the determining factor for a solid performance. To analyze the performance of Gemma 2 9B further, we look at the distribution of the labels in Figure~\ref{fig:fig3} concerning the ground truth labels presented in Figure~\ref{fig:fig2}. It can be deduced that the model is better at predicting the negative and neutral sentiment than at predicting the positive sentiment. Evaluating all the LLMs in this way we found out that this is also the case for all of the others. The models detect much more negative stance polarity than positive ones and when looking at the distribution of the ground truth labels in \ref{fig:fig2} it can be seen that some positive labels are predicted as negative by the models. In the context of the decision conferencing simulation system, this is not that bad as it would be worse if the negative stance polarity would not be detected accurately, leading to an earlier termination of the debate of the participants even though no real agreement is reached.

The evaluation of stance polarity detection on the Claim Stance Classification dataset reveals important insights into the capabilities of modern LLMs. Only one of the six evaluated models was able to surpass the baseline sentiment accuracy, which indicates that stance polarity detection remains a challenging task, particularly in datasets with imbalanced classes. Considering the $F_1$~scores of the evaluated LLMs contained in Table~\ref{table:4} it becomes clear that LLMs still have some weaknesses in predicting the stance polarity on the Claim Stance Classification dataset. But the results indicate that using the polarity stance might in some cases help to indicate especially the negative sentiment and enrich the agreement detection with this factor, although the detection of agreement with LLMs should not rely solely on the detection of sentiment in claims related to some topic.

In conclusion, the objective evaluation of six different LLMs across various datasets for stance detection and stance polarity detection demonstrates that LLMs are highly effective, particularly due to their zero-shot capabilities. These models often outperform task-specific, fine-tuned models and consistently deliver strong results without the need for extensive training. Our findings also highlight that it is not necessary to rely on the largest or closed-source models to achieve competitive performance as Gemma 2 9B and LLaMA 3 70B are also proficient in the different tasks. This suggests that mid-sized models can offer a balance of efficiency and accuracy, making them practical choices for the application in our decision-conferencing system. The application of these models as a judge agent in our proposed decision conference system is evaluated in the next section to find out if the results of the objective evaluation transfer to an application in a real-world scenario. 

\subsection{Subjective Evaluation of Large Language Models with a Simulated Decision Conference}
To begin the subjective evaluation, we focus on the ability of our decision conferencing system introduced in section \ref{sec3} to detect agreement among debating participant agents. This assessment centres on qualitative measures, exploring how well the system supports collaboration and decision-making in a simulated decision conference. Specifically, we evaluate the role of a judge agent, which is tasked with detecting agreement between participants, using a combination of the LLM-as-a-judge approach for the effectiveness of the judge agent and a manual evaluation of the system with and without the judge agent.

The evaluation is based on an example decision conference explained in section \ref{sec4}, which is simulated using the decision conference system proposed in \ref{sec3}. By analyzing the system’s performance in this context, we can gain useful insights into its effectiveness in fostering cooperation and its potential for enhancing group decision-making processes. This subjective evaluation adds another dimension to the objective evaluation of the LLMs by addressing the system's real-world usability and its capacity to manage complex interactions.

\subsubsection{Evaluation of the Detection of Agreement in Decision Conferences}
With the general procedure for the subjective evaluation explained in section \ref{sec4}, we first evaluate the capabilities of each of the six LLMs in agreement detection in simulated decision conferences. Since the simulated decision conference explained in section \ref{sec3} incorporated three topics to debate, every LLM has several chances to detect agreement and decide if the participants must debate further or if they should move to the next topic. To keep the prompts for the LLM acting as an LLM-as-a-judge concise to not exceed any context windows, we split the debates based on each decision made by the judge agent. For each evaluation, only one decision is considered at a time. We select five decisions from the judge agent for evaluation, applying this to each of the six models. 

Since a strong LLM is needed to fulfil the needs of the LLM-as-a-judge approach we decided to use ChatGPT 4 as the model evaluating the models on the task of agreement detection in decision conferences \cite{zheng2024judging}. Using ChatGPT 4 as an evaluator helps to increase the consistency and reliability of the subjective evaluation, as its advanced reasoning and language understanding capabilities shown in many benchmarks since its release, allow it to effectively evaluate the assessments of other models and it also has the highest agreement with human evaluators on different tasks \cite{zheng2024judging}. The criteria are the evaluation of the decision, namely whether the model correctly decides whether agreement was reached or not and whether the answer is formatted correctly. By isolating the decision points and evaluating them individually, we ensure that the models are assessed in a focused and controlled manner for their ability to recognise agreement. This approach minimises the possibility of external factors influencing the assessment and provides a better insight into the strengths and weaknesses of each model in this particular task.

Table~\ref{table:5} presents the scores for each part of the debate, as evaluated by the LLM-as-a-judge approach on a scale of one to ten, along with an overall score for each model. It can be seen that the three best models from the objective evaluation, Llama 3 70B, Gemma 2 9B and ChatGPT 4, remain the top performers in the subjective evaluation. Their scores are also very close to one another, with a significant gap separating them from the other models' scores.

\begin{table}[t]
    \caption{Evaluation of Agreement Detection using LLM-as-a-judge-approach on Simulated Decision Conferences with scores between 1 and 10}
    \label{table:5}
    \setlength{\tabcolsep}{3pt}
    \begin{tabular}{ lrrrrrr}
        \toprule
       Model & Debate 1 & Debate 2 & Debate 3 & Debate 4 & Debate 5 & Overall\\
        \midrule
        Gemma 2 9B & 8 & 10 & 10 & 10 & 10 & \textbf{9.6}\\
        Mixtral 8x7B & 2 & 1 & 1 & 2 & 4 & 2\\
        Gemma 7B & 10 & 2 & 2 & 4 & 1 & 3.8\\
        Llama 3 70B & 8 & 10 & 10 & 10 & 10 & \textbf{9.6}\\
        ChatGPT 3.5 Turbo & 3 & 1 & 10 & 10 & 3 & 5.4\\
        ChatGPT 4 & 10 & 10 & 10 & 10 & 10 & \textbf{10}\\
        \bottomrule
    \end{tabular}
\end{table}

This result highlights a strong relationship between the performance on objective metrics and the models' performance on subjective metrics for tasks like agreement detection. The closeness of the top models' results suggests that high-performing LLMs tend to perform well not only on the benchmark datasets but also on complex decision scenarios that require a deeper understanding of context, discussion structure and interpersonal agreement dynamics. The clear gap between the top-performing models and the remaining models emphasises the different levels of competence of LLMs in tackling such challenging tasks and also the ability of the different models to answer in the correct format needed for the rest of the system to work correctly.

It is particularly interesting that despite its significantly smaller size, Gemma 2 9B demonstrates strong agreement detection capabilities during the simulated decision conferences, closely matching the performance of ChatGPT 4. This demonstrates that it can also be sufficient to use a smaller model for this task, also keeping in mind the requirements to run this model are not as high and therefore the environmental impact is also less. The same applies to the Llama 3 70B model since it is still reasonably sized. 

This subjective evaluation of six different LLMs using the LLM-as-a-judge shows that the models are capable of detecting agreement in the debates of participants in a decision conference. It not only confirms the top models established by objective metrics but also further confirms the assumption that task-specific performance may not always correlate directly with model size. This finding has important implications for the development of LLMs, particularly in resource-efficient applications. 

\subsubsection{Comparison of the System with and without Agreement Detection}
Having demonstrated the system's ability to recognise agreement by the judge agent in a simulated decision conference, we now move on to an analysis of the system's performance with and without agreement detection enabled. This comparison provides information on how agreement detection affects the overall decision-making process and how it influences the collaboration of participants, the speed of consensus building and the quality of decisions made. By evaluating both setups, we can better understand the added benefit of agreement detection and its ability to improve the effectiveness of the system in real-life scenarios.

For this evaluation we examine the output of two simulated decision conferences, using the real decision conference presented in section \ref{sec3} as an example. The first simulation is done without the judge agent and just comprises two participants and a moderator, while the second simulation adds a judge agent. Since there are again three topics to discuss, we look at the results of the debate on the first topic and compare it to the results of the real decision conference to get an idea of which systems help us to get closer to the real outcome.

As explained in section \ref{sec3}, the first topic discussed in the example decision conference from \cite{rogeberg2018new} is the development of a set of criteria for assessing drug policy outcomes. In the real decision conference, the experts concluded that policies should be evaluated based on 27 criteria, grouped into seven thematic clusters: health, social, political, public, crime, economic, and cost \cite{rogeberg2018new}.

In the simulated decision conference without a judge agent, the participant agents manage to cover six of the clusters, but the public cluster is not addressed before the moderator moves on to the next topic. The transcript of the debate on this topic can be viewed in Figure~\ref{fig:fig4}. This highlights an important issue: the debate progresses too quickly, without fully exploring each aspect of the topic. This can be a big problem if the requirement for the outcome of the decision conference is an answer that highlights every possibility and takes into account all kinds of different perspectives and the moderator agent does not seem to catch it at all times. 

In the simulation with a judge agent, the agent determined that the debate was not complete after one round of discussion among the participants. As a result, the previously overlooked public implications cluster is addressed, allowing the simulated decision conference to reach the same outcome as the real one. The debate with the judge agent involved is displayed in figures \ref{fig:fig5} and \ref{fig:fig6}. 

This short evaluation of only one debate still demonstrates the value of including a specialized agent to detect agreement among participants in a simulated decision conference. The judge agent not only enhances the depth of the debates but also assists the moderator in determining the appropriate time to transition to the next topic, ensuring more thorough and balanced discussions. By ensuring that no important aspect of a debate is left unexplored, the judge agent helps prevent premature transitions between topics and improves the overall quality of the discourse. This method allows for more comprehensive coverage of each topic, leading to outcomes that are closer to real-world decision-making processes. This approach can help to ensure that the quality of debate remains high and that all relevant perspectives are considered.

\section{Conclusion}\label{sec6}
In this work, we developed a system to simulate decision conferences using LLMs, modelling the structure and processes of real-world decision conferences. We conducted both objective and subjective evaluations of the LLM judge agent, which is responsible for detecting agreement among participants. Through these evaluations, we highlighted the critical role agreement detection plays in debates within simulated decision conferences and addressed the research questions outlined in Section \ref{sec1}. Our results demonstrated that LLMs can be highly effective in agreement detection, outperforming some models specifically trained for stance detection and stance polarity detection on benchmark datasets. Interestingly, smaller and mid-sized models performed as well as, or even better than, the standard ChatGPT 4 model. This finding has important implications for future research and development, suggesting that task-specific applications could benefit from the application of smaller models. The short subjective evaluation confirmed that LLMs can be used to detect agreement in debates where participants hold divergent perspectives and differences in opinion are actively sought. But this must also be further explored using more decision conferences to ensure the results are applicable for all scenarios, even though it is difficult to find real decision conferences with the results available. Models such as Gemma 2 9B and LLaMA 3 70B performed comparably to ChatGPT 4 in real-world scenarios. Since these models are open-source with permissive licenses, they offer promising opportunities for further research, including fine-tuning for specific applications. 

Using LLMs as judge agents for agreement detection in simulated decision conferences also presents several challenges. While the flexibility of LLMs to adapt to different topics reduces the need for specialized training or fine-tuning, their accuracy across all topics remains uncertain. For certain subjects, it may still be advantageous to train a classical model rather than rely solely on an LLM. Another limitation we observed is the tendency of LLMs to overlook parts of the prompt, a persistent issue even with newer models. Since the system relies on the judge agent's correct interpretation of the prompt, some LLMs proved unreliable, and even ChatGPT 4 occasionally struggled with following exact instructions. We did not explore the potential of prompt engineering to address this issue in-depth, so this remains an area for future investigation. Prompt robustness and hallucinations also pose challenges to the entire system. While the judge agent effectively detects agreement, it does not evaluate whether the points made by the participant agents are accurate or relevant to the debate. This is a well-known limitation of LLMs, and future research could explore techniques like retrieval-augmented generation (RAG) or knowledge graphs (KG) to ground agents in more relevant information, ensuring their contributions are aligned with the topic at hand.

These approaches could also mitigate another issue we encountered: LLMs sometimes possess knowledge that is either too advanced or insufficient for the topic being discussed. Ideally, the behaviour of participant agents should mirror human participants, but if an LLM’s knowledge far exceeds that of a typical human, the debate process may be short-circuited, with agents reaching an agreement too quickly without meaningful discussion. Constraining the knowledge of LLM agents, possibly through RAG or KGs, could prevent this and allow for a more natural flow of debate, leading to richer insights.

In summary, our evaluation of agreement detection within the decision conferencing simulation system highlights its importance for facilitating debate in LLM-based multi-agent systems. Our method, which involves using a judge agent to identify agreement, could enhance systems that employ internal debate for decision-making across different tasks. Therefore, this study presents an opportunity for further exploration of the application of these techniques in diverse collaborative decision-making scenarios ultimately enhancing both the practical capabilities and theoretical understanding of such systems.






\section{Compliance with Ethical Standards}
\subsection{Declarations}

\subsubsection{Funding}
No funds, grants, or other support was received.

\subsubsection{Conflict of interest}
The authors have no financial or proprietary interests in any material discussed in this article.

\subsubsection{Ethics approval and consent to participate}
Not applicable

\subsubsection{Consent for publication}
Not applicable

\subsubsection{Data availability}
The varied stance topic (VAST) dataset from \cite{allaway-mckeown-2020-zero} is available at \url{https://github.com/emilyallaway/zero-shot-stance} and the claim stance classification from \cite{bar-haim-etal-2017-stance} is available at \url{https://research.ibm.com/haifa/dept/vst/debating_data.shtml#Argument_Stance} as the ``IBM Debater - Claim Stance Dataset".

\subsubsection{Materials availability}
Not applicable

\subsubsection{Code availability}
Not applicable

\subsubsection{Author contribution}
Not applicable

\begin{appendices}

\section{Transcripts of the Experiments with Simulated Decision Conferences}\label{secA1}

\begin{figure}[t]
    \centering
    \includegraphics[width=0.8\textwidth]{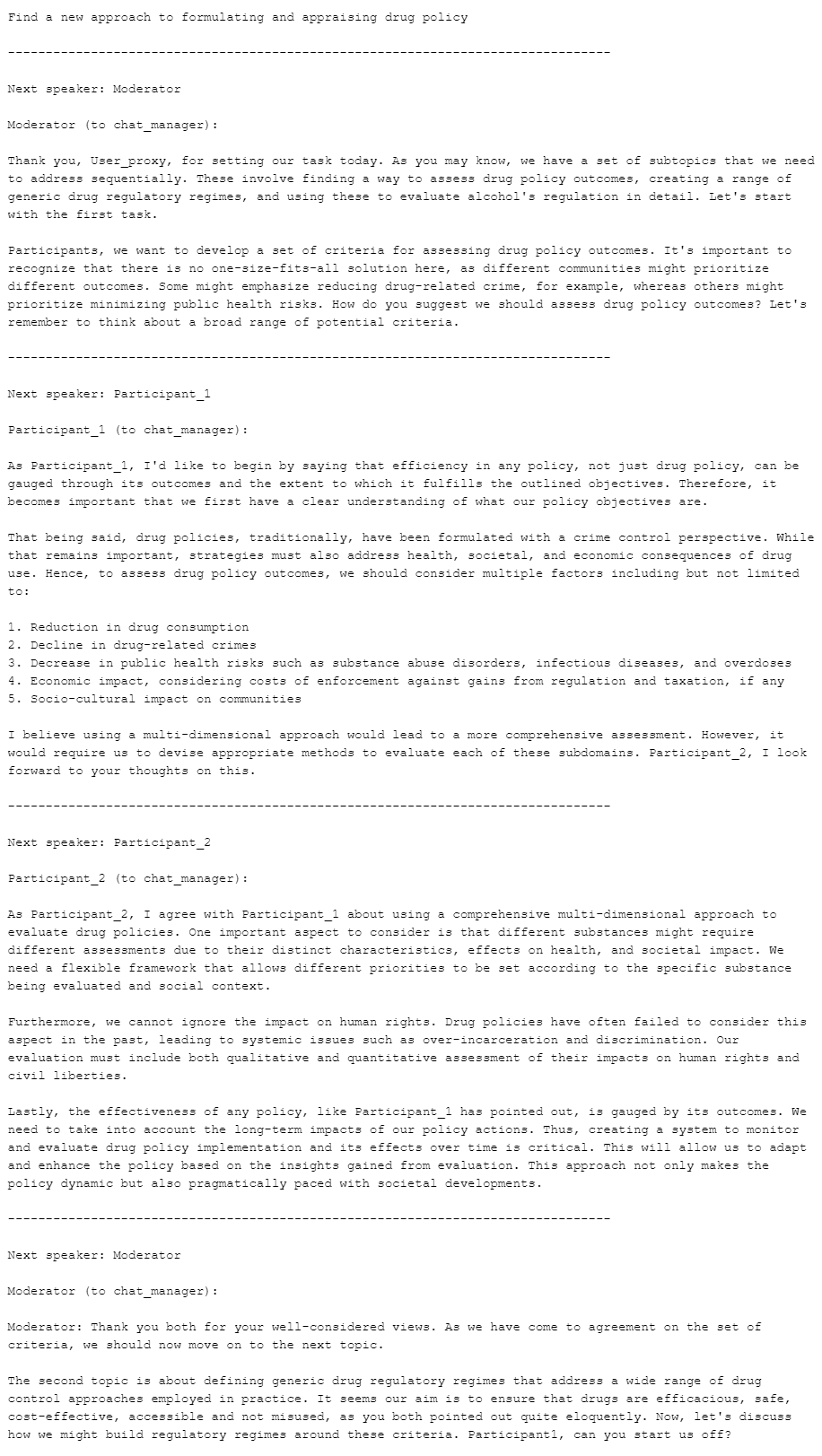}
    \caption{\label{fig:fig4}Output of the transcript of simulated decision conference for the first topic without the judge agent}
\end{figure}

\begin{figure}[t]
    \centering
    \includegraphics[width=0.85\textwidth]{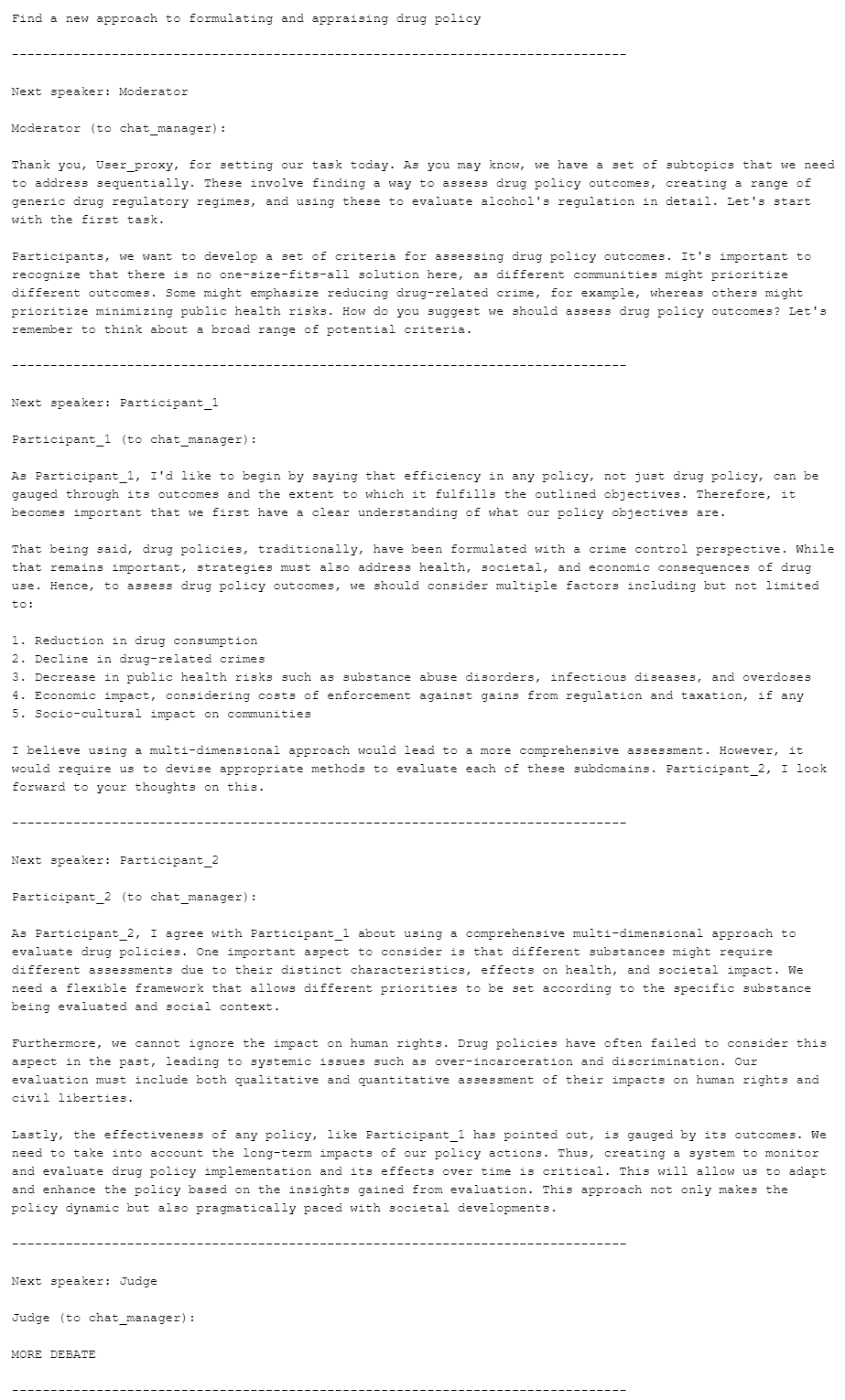}
    \caption{\label{fig:fig5}Part 1 of the transcript of simulated decision conference for the first topic with the judge agent}
\end{figure}

\begin{figure}[t]
    \centering
    \includegraphics[width=\textwidth]{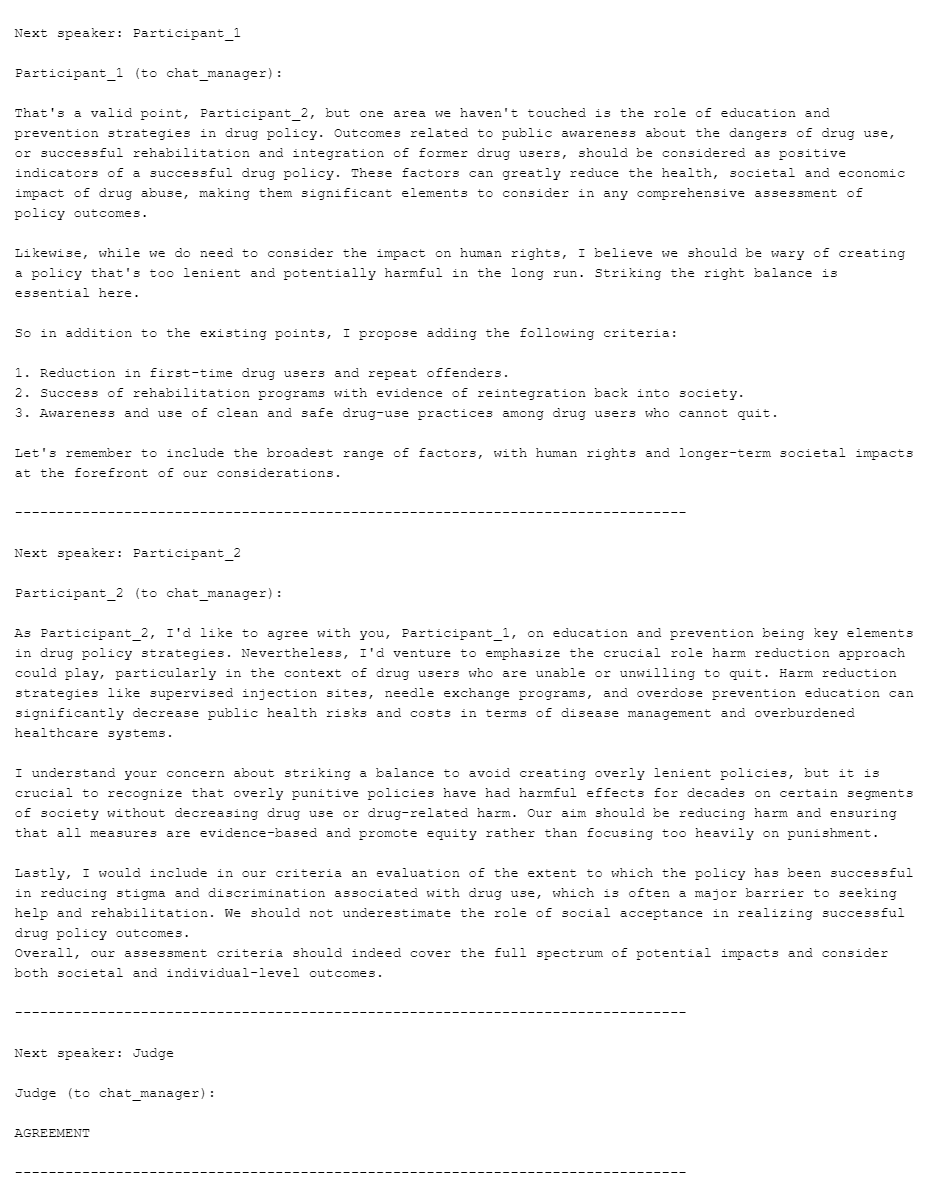}
    \caption{\label{fig:fig6}Part 2 of the transcript of simulated decision conference for the first topic with the judge agent}
\end{figure}




\end{appendices}


\bibliography{sn-bibliography}

\end{document}